\def\year{2022}\relax
\documentclass[letterpaper]{article} 
\usepackage{aaai22}  
\usepackage{times}  
\usepackage{helvet}  
\usepackage{courier}  
\usepackage[hyphens]{url}  
\usepackage{graphicx} 
\usepackage{verbatim}
\usepackage{natbib}  
\usepackage{caption} 
\DeclareCaptionStyle{ruled}{labelfont=normalfont,labelsep=colon,strut=off} 
\usepackage{algorithm}
\usepackage{algorithmic}

\usepackage{newfloat}
\usepackage{listings}
\floatstyle{ruled}
\newfloat{listing}{tb}{lst}{}
\floatname{listing}{Listing}
\title{Classifying Emails into Human vs Machine Category}
\author{
    Changsung Kang, \textsuperscript{\rm 1}
    Hongwei Shang, \textsuperscript{\rm 2}
    Jean-Marc Langlois \textsuperscript{\rm 2}

}
\affiliations{

    \textsuperscript{\rm 1} Walmart Global Tech \\
    \textsuperscript{\rm 2} Yahoo Research \\
    changsung.kang@walmart.com, hongwei\_shang@yahoo.com, jmlanglois@yahoo.com

}

\usepackage{bibentry}

\begin{document}

\maketitle

\begin{abstract}
It is an essential product requirement of Yahoo Mail to distinguish between personal and machine-generated emails.
The old production classifier in Yahoo Mail was based on a simple logistic regression model. That model was trained by aggregating 
features at the SMTP address level.
We propose building deep learning models at the message level.
We built and trained four individual CNN models:
(1) a \textit{content model} with subject and content as input;
(2) a \textit{sender model} with sender email address and name as input;
(3) an \textit{action model} by analyzing email recipients' action patterns and
correspondingly generating target labels based on senders' opening/deleting behaviors;
(4) a \textit{salutation model} by utilizing senders' ``explicit salutation'' signal as positive labels.
Next, we built a final \textit{full model} after exploring different combinations of the above four models.
Experimental results on editorial data show that our full model improves the adjusted-recall from 70.5\% to 78.8\% compared to the old production model, while at the same time lifts the precision from 94.7\% to 96.0\%.
Our full model also significantly beats the state-of-the-art Bert model at this task.
This full model has been deployed into the current production system (Yahoo Mail 6).
\end{abstract}

\section{Introduction}
\label{sec:intro}

\citet{Grbo:Hala:how:2014} showed that today's mail web traffic is dominated by machine-generated messages, originating from mass senders, such as social networks, e-commerce sites, travel operators etc; and a classifier should first distinguish between machine and human-generated messages before attempting any finer classification. A human generated email comes from an individual person. A machine generated email is an automatic email, a mass marketing email, and/or a template email. The annotation guideline details about Human/Machine email's definition are in Appendix.

There are a lot of well-known applications on mail area, 
including spam detection \citep{Kuma:Sara:Visu:2019,douzi2020hybrid,mohammadzadeh2021novel}
user's future action prediction \citep{Cast:Zoha:You:2016}, email threading \citep{Ailo:Karn:Thre:2013} and 
information extraction \citep{Agar:Jite:Temp:2018,Di:Ifta:Auto:2018,Shen:Tata:Anat:2018}.
Recently, research community has made more interesting applications,
such as cyber security events detection \citep{Vina:Soma:Scal:2019},
commitment detection \citep{Azar:Sim:Doma:2019}, intent detection \citep{10.1145/3397271.3401121} and online template induction \citep{Whit:Edmo:Onli:2019}.
Most of these applications depend on training high-quality classifiers on emails. For mail classification, there has been some research efforts to learn email embedding by leveraging various aspects/features of emails.
\citet{Sun:Garc:Lear:2018} used only the sequence of B2C templates as input to learn representations. 
\citet{Koca:Shen:Rise:2019} utilized HTML structure of emails to help classify machine-generated emails.
\citet{Pott:Wend:Hidd:2018} leveraged both text and image content of commercial emails to classify emails. 

The recent advancement in natural language processing (NLP), especially in text classification, has proposed various deep learning model architectures for learning text representation in place of feature engineering.
Fundamental work in this area involves with learning word representations by predicting a word's context \citep{Miko:Suts:Dist:2013,Boja:Grav:Enri:2017}. Later work extended to learn representations for
longer sequences of text, at sentence-length level and document-length level \citep{Yang:Yang:Hier:2016,Le:Miko:Dist:2014,Kiro:Zhu:Skip:2015,Joul:Grav:Bag:2016}.
Most recently, motivated by the Transformer architecture \citep{Vasw:Shaz:Atte:2017}, the state-of-the-art BERT model \citep{Devl:Chan:Bert:2018} and its variations \citep{Yang:Dai:XLNe:2019,Liu:Ott:Robe:2019,lan2019albert} pre-trained bidirectional representations from unlabeled text.

It is an essential product requirement of Yahoo Mail to distinguish between personal and machine-generated emails. The model proposed by Yahoo researchers \citep{Grbo:Hala:how:2014} has been deployed into previous Yahoo Mail production system. 
This model relies on a large number of well designed features for prediction,
including content, sender, behavioral, and temporal behavioral features. 
In spite of its accurate prediction, this old production model has a ``sender cold-start'' issue with 70\% of email traffic covered \citep{Grbo:Hala:how:2014}.
The model may not be able to conduct appropriate learning for new senders with few associated messages, especially 
when the most predictive features are historical behavioral features and 
new users do not have enough history to provide reliable features. 
Another drawback of the production model is that it is a simple logistic regression model without utilizing the recent development in NLP. In addition to considering behavior features, the previous production model uses only bag of words as content input 
without taking into account signals from text sequence like long short-term memory (LSTM) and CNN models.

In this work, we propose to classify messages into
human/machine category by solely relying on the message itself as input
without any additional feature engineering. 
To further improve prediction accuracy,
useful user behavior signals are utilized as target labels rather than input:
separate training data sets are created by generating target labels from these signals, 
and individual sub-models are built by keeping the same input for each set of created training data.   
Finally, the full model combines these sub-models for final prediction.
By directly building the full model at message level, 
we naturally solve the cold-start issue as only message
itself can predict well without any behavioral information as input. 

In summary, we are not aware of any work at mail classification that 
(1) builds CNN models based on generated target labels from users' behavior signals 
-- senders' explicit salutation at the 
beginning of message and recipients' opening/deleting behaviors;
(2) explores potential options for combining sub-models into a final model and analyzes it. 
We aim to deliver a human/machine classifier
that answers the following questions:
\begin{itemize}\setlength\itemsep{-0.3em}
    \item \textit{Can we automatically learn representations from the message itself without 
    any feature engineering in order to predict the class?} We show that the \textit{content model} with email subject and content only as input can achieve better performance than the feature-engineered 
    logistic regression model with 2 million features
    (Section~\ref{sec:experiment}).
    \item \textit{Can we leverage some useful behavioral signal to further improve the model performance? 
    How can these signals help the model without bringing burden of saving/updating extra feature inputs? } 
    Instead of adding these signals as feature inputs, 
    we trained two individual CNN models by generating target labels from users' behaviors: 
    one based on recipients' opening/deleting behaviors, and the other based on senders' ``explicit salutation'' at the beginning of email. 
    Then our full model combined these models into one final classifier.
    Specifically, the proposed final \textit{full model} improves the adjusted-recall by 11.8\% compared against the old production model, while at the same time lifts the precision from 94.7\% to 96.0\%. 
    \item \textit{Can we propose a model with great prediction accuracy based on pure message information, with simple model structure and good inference latency in order to deploy it on production system?} 
    We keep a simple neural network model structure (Section~\ref{sec:model}), and build our own vocabulary dictionary. 
    With comparable number of words as GloVe 
    , models built with our own dictionary can achieve the same prediction accuracy with embedding dimension 64,
    in contrast to using GloVe word embedding with dimension 300.
    This reduces our model size by almost 4 times smaller than using pre-trained GloVe embedding.
    The full model is deployed into Yahoo Mail 6 production system, 
    and is able to finish running inference on 30M messages within one hour. 
\end{itemize}

The rest of this paper is organized as follows. 
We introduce our proposed model architecture in Section~\ref{sec:model}.
We present details on how training and testing data are obtained in Section~\ref{sec:data}.
Our experiments and results are shown in Section~\ref{sec:experiment},
with adjusted precision/recall computation presented in the Appendix.
Section~\ref{sec:conclude} concludes the ongoing work.

\section{Model Architecture}
\label{sec:model}

In this section, we introduce the design of the model architecture for human/machine classifier. 
We first introduce the individual sub-models based on different features and signals separately. 
For all sub-models, the main component is the temporal convolutional module, 
which simply computes a 1-D convolution using temporal convolution, batch 
normalization, and max-pooling to deal with the temporal textual data.
The final full model combines the four CNN models by joining each model's individual representations at different levels.

Before exploring these model structures, let us first define some notations.

\paragraph{Convolutional Block}
First, for the ease of the notation, we make use of convolutional block, following \citet{Ediz:Mant:Deep:2017}. 
The temporal convolutional module consists of a set of filters with various sliding window sizes. 
Each filter applies a convolution operation on its input, with its weights learnt during the training.
By running a contextual sliding window over the input word/letter-trigram sequence, each filter can learn
its pattern. In this work, we use a stride of 1. We represent our 
convolutional block  by ``\textit{Temp Conv,[1,2,...,k],f}'', where [1,2,...,k] represents all sliding 
window sizes applied for convolution operation and $f$ corresponds to the number of filters for each sliding window size in [1,2,...,k].
Max pooling is used to force the model to keep only the most useful local features produced by the convolutional layers;
the maximum value across each sliding window over the whole sequence is selected for each filter. 
Batch normalization is also used to normalize its input, which brings an additional regularization effect and
as a result, accelerates the training \citep{Ioff:Szeg:batc:2015}.

\begin{figure}[!t]  
  \centering
  \includegraphics[width=0.45\textwidth]{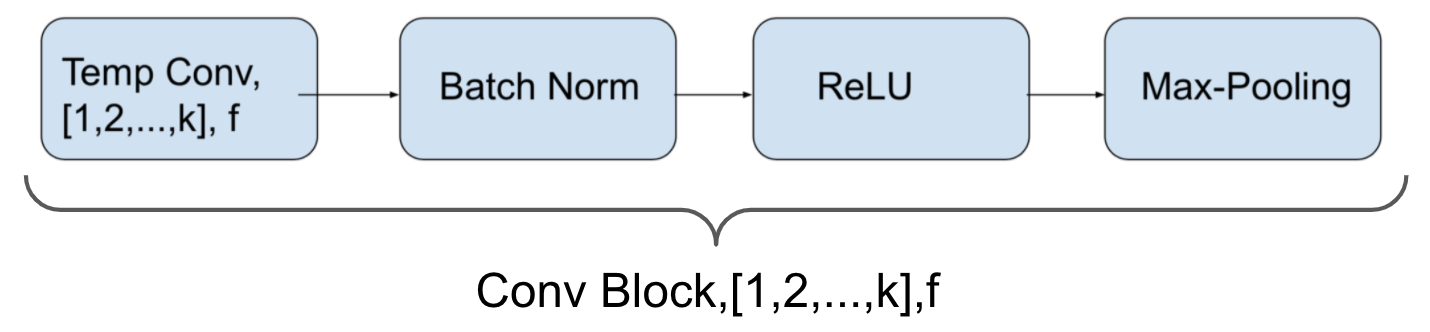}
  \caption{Detailed illustration of a convolutional block.}
  \label{fig:convBlock}
\end{figure}

A convolutional block (see Figure~\ref{fig:convBlock}) consists of a sequence of one convolutional layer,
a batch normalization layer, a ReLU activation function, and finally a max-pooling layer;
it will be denoted by \textit{Conv Block,[1,2,...,k],f} with the number of filters f for each sliding 
window size in [1,2,...,k].

\paragraph{Other functions}
In addition to ``\textit{Conv Block,[1,2,...,k],f}'', we denote fully connected layers
with no activation function by ``\textit{FC, O, L1, L2}'' where O is the number of output neurons
and L1, L2 represents the penalty parameters for L-1 and L-2 regularization respectively.
For simplicity, we use ``\textit{BN, ReLU}'' to denote a batch normalization layer followed by a ReLU
activation function. 

In all of our models, we fix the sequence length at $s$, with subscripts used to denote for each
specific feature. For example, $s_{subject}$ represents the sequence length for the subject input.
When the sequence length is smaller than $s$, the remaining positions are fully padded with index 0. 
When the sequence length is larger than $s$,
we simply ignore all the words appearing after the s\textsubscript{th} position of the sequence.

Each sub-model considers a word-unigram/letter-trigram dictionary, 
and creates a vectorial representation for each word or letter-trigram, depending on the specific model. 
We use $V$ with subscripts to denote dictionary for different features, with $V_{w}$, $V_{trig}$, $V_{name}$
corresponding to word dictionary for subject and content, letter-trigram dictionary for sender email address,
and word dictionary for sender name respectively. 

Each single model is detailed in the following.

\subsection{Content Model}
\label{sec:content_model}
\begin{figure}[!t]  
  \centering
  \includegraphics[width=0.45\textwidth]{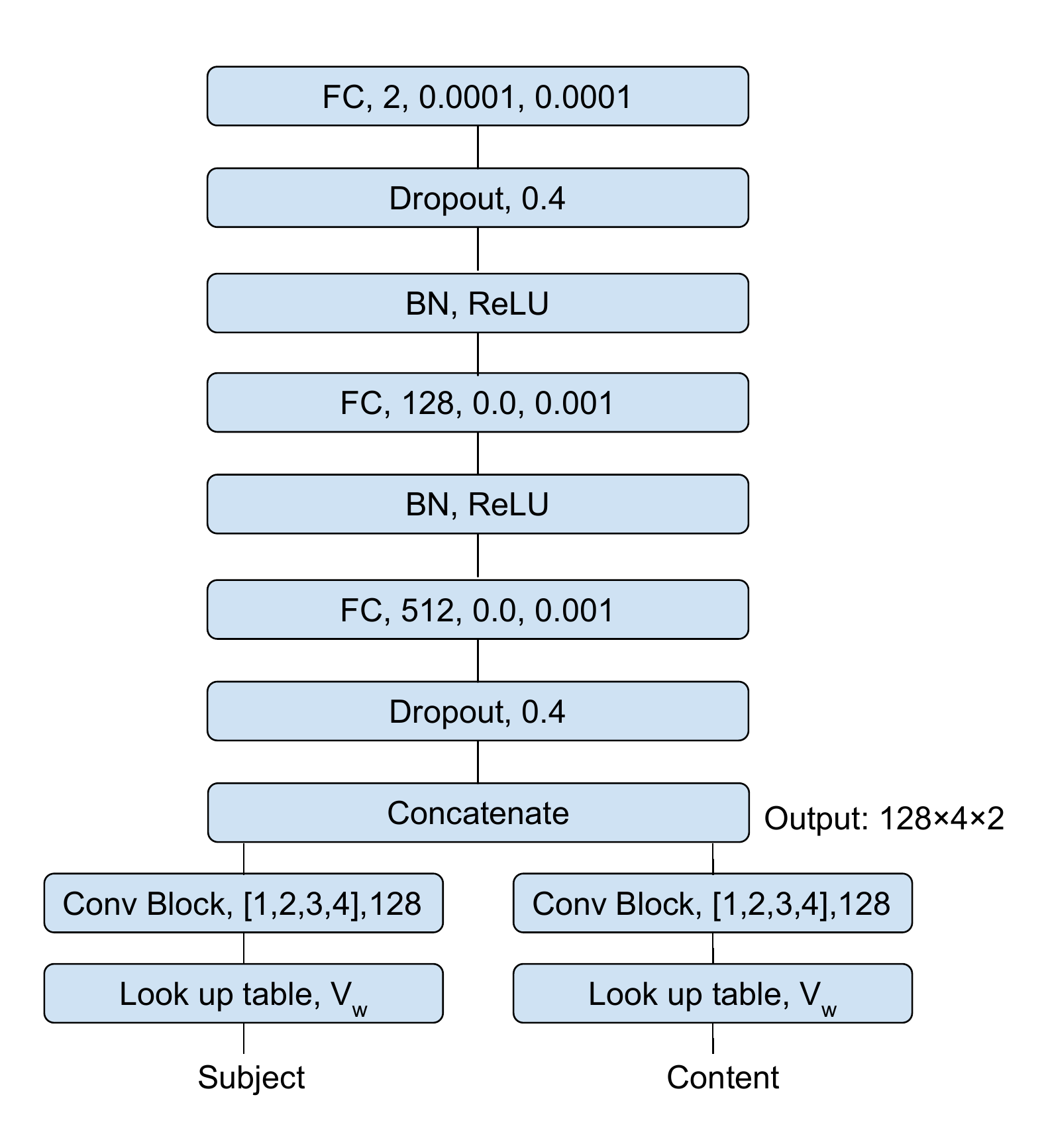}
  \caption{Model structure for content model and action model.}
  \label{fig:basemodel}
\end{figure}

The content model (also called base model in this manuscript) uses email subject and content as inputs to predict human/machine (H/M) category,
with the model structure shown in Figure~\ref{fig:basemodel}. 

In this base model, the inputs are a fixed-size padded subject with sequence length $s_{subject}$
and a fixed-size padded content with sequence length $s_{content}$. 
This model starts with a look-up table that creates a vectorial representation for each word and converts both subject and content inputs into
tensors with size $s_{subject} \times e$ and $s_{content} \times e$ separately, where $e$ is embedding dimension size. 
We first apply two separate convolutional blocks ``\textit{Conv Block,[1,2,...,k],f}'' with $k=4$ and $f=128$ 
for both subject and content inputs,   
resulting in a one-dimensional tensor with length $k \times f=512$ for both subject and content inputs.
These two tensors are concatenated into a single vector with length 1024, which then goes through a dropout layer with dropout rate $r=0.4$. 
The resulting vector will be the input to two fully connected
layers, with each fully connected layer followed by a batch normalization and ReLU activation function.
After using one more dropout with the same dropout rate, a final layer with two hidden units is 
implemented before applying softmax function to calculate estimated probability score.

\subsection{Sender Model}

\begin{figure}[!t]  
  \centering
  \includegraphics[width=0.45\textwidth]{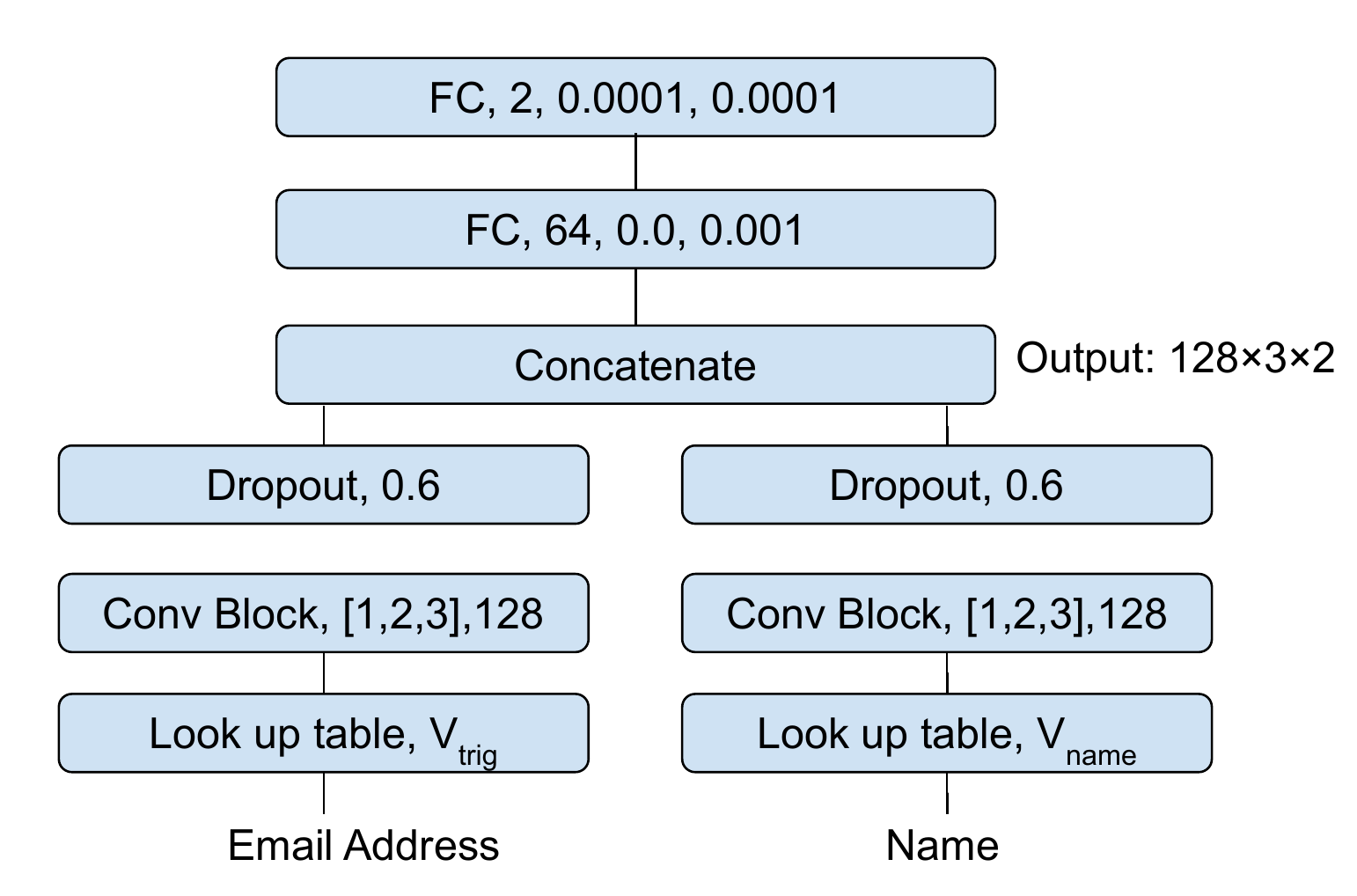}
  \caption{Model structure for sender model.}
  \label{fig:sendermodel}
\end{figure}

The sender model uses sender's email address and name as inputs to predict H/M category,
with the model structure in Figure~\ref{fig:sendermodel}. 

The model contains 
(1) a letter-trigram representation layer obtained by running a contextual sliding window over the input sequence on sender email address 
followed by creating vectorial representation for each letter-trigram through V\textsubscript{trig};
(2) in parallel with the above step, a word-unigram representation layer obtained by creating vectorial representation for each word in sender name through V\textsubscript{name};
(3) a convolutional block \textit{Conv Block,[1,2,3],128} and a followed dropout layer with dropout rate 0.6 for both resulting representations
obtained from sender email address and name input;
(4) a concatenation layer that concatenates the resulting sender email features and name features obtained from the previous step;
(5) a fully connected layer with 64 output neurons and L2-norm penalty with penalty parameter 0.001;
(6) a final fully connected layer that output 2 units, containing both L1-norm and L2-norm with the same penalty parameter 0.0001. 

\subsection{Action Model}
To leverage user behaviors from different sources, 
we study the relationship between an email's H/M category and the email recipient's action with the question ``how an email recipient's action is affected by the email's H/M category?''.
We focus on ``open'' and ``delete'' actions with the assumption that human messages tend to be opened more, and deleted
less than machine messages do. Table~\ref{tab:action_explore} shows the percentages of messages for human, machine, and unknown
categories under different recipients' actions from labeled data. 
With random sampling from the whole email pool, 
5.10\% of sampled messages belong to human category.
Restricting the sampling based on deleted and not opened messages (denoted by $B$)
will produce only 1.74\% human category messages. 
Restricting the sampling from messages that are opened and not deleted by recipients (denoted by $A$)
will produce 26.94\% human category messages.
A higher percentage, 75.23\%, of human category messages can be obtained by further filtering set $A$ (denoted by $A \setminus B$):
removing messages whose corresponding senders appear in $B$ from $A$. 
Note that the above statistics are calculated by taking into account more than 20\% messages in
unknown category. 
Ignoring the unknown messages for action $A \setminus B$ will yield to 96.25\% of human messages.

\begin{table}
\centering
\small
\begin{tabular}{rrrrr}
\hline
 & random & open, not  & not open,  & A$\setminus$B \\
 &        & deleted $(A)$ & deleted $(B)$ &  \\
\hline
 H &  5.10 & 26.94 & 1.74 & 75.23 \\
 M  & 67.02 & 49.31 & 76.19 & 2.93  \\
 U &  27.88 & 23.75 & 22.07 & 21.84  \\
\hline
\end{tabular}
\caption{Percentage of messages for human (H), machine (M), and unknown (U) categories respectively under different actions.}
\label{tab:action_explore}
\end{table}

Based on the above strong correlation between message's category and the recipients' action on messages,
we generate a set of training data with
\begin{itemize}\setlength\itemsep{-0.3em}
    \item messages that are ``deleted and not opened'' ($B$) as negative label 0;
    \item messages that are ``open and not deleted'' with removing senders appearing in $B$ and corresponding messages ($A \setminus B$) as positive label 1.
\end{itemize}

With this newly generated message-level training data, we train a CNN model with subject and content as input, using the same model structure as content model (see Figure~\ref{fig:basemodel}).

\subsection{Salutation Model}

This model is motivated by analyzing false negative messages produced by the previous production model. 
We find out that messages with explicit salutation (e.g. starting with ``Dear ...'') 
tend to be sent by someone who knows the person, thus belonging to human messages.
Here, a message is defined to contain ``explicit salutation'' if
there is a match between the recipient names and the beginning part of the email body.
Considering this relationship, we generate another set of training data with 
\begin{itemize}\setlength\itemsep{-0.3em}
    \item messages that contain explicit salutation as positive label 1;
    \item messages that do not contain explicit salutation as negative label 0.
\end{itemize}

\begin{figure}[!t]  
  \centering
  \includegraphics[width=0.45\textwidth]{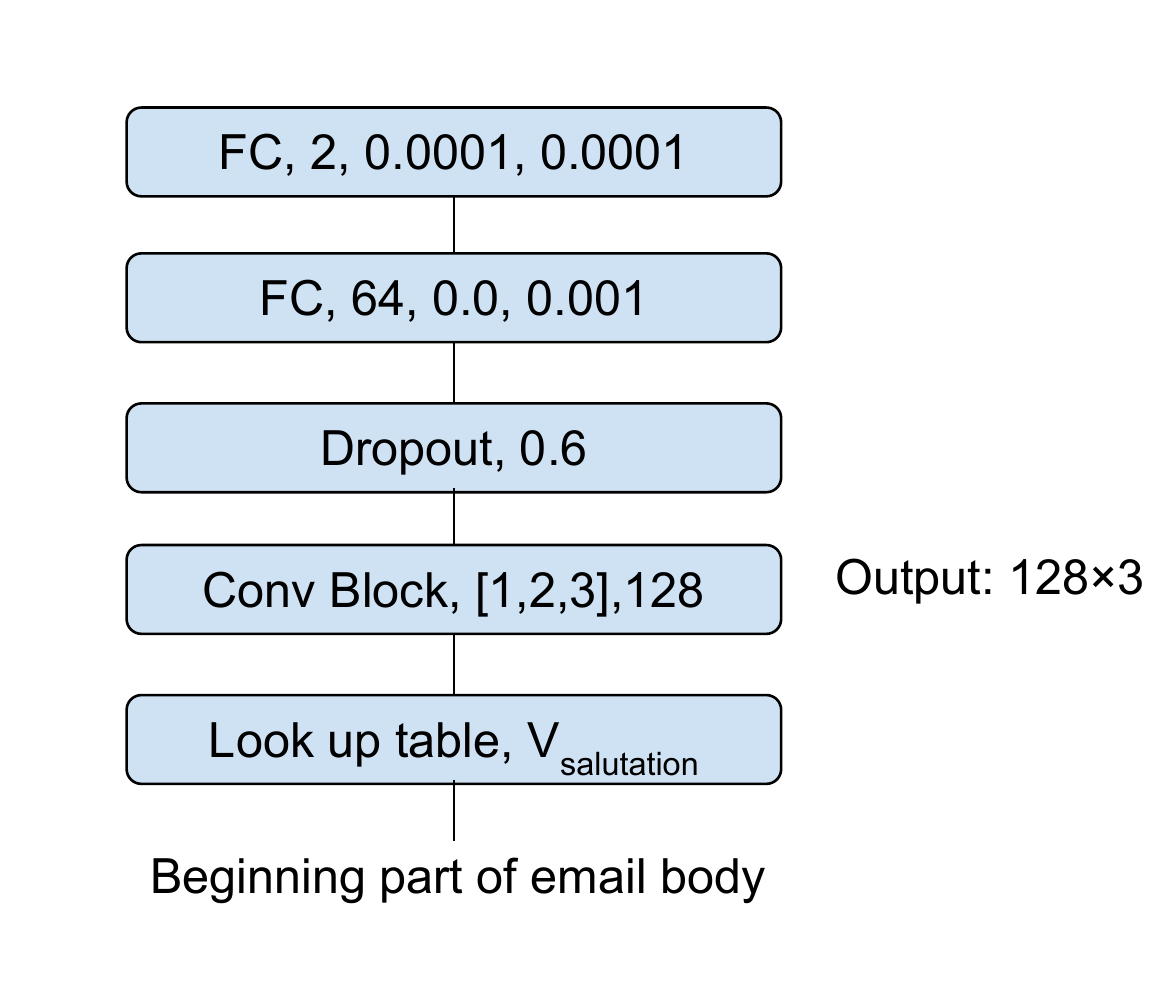}
  \caption{Model structure for salutation model.}
  \label{fig:salutationmodel}
\end{figure}

With the generated message-level training data, we train a CNN model
using the beginning part of email body as input, 
which is defined as the sequence of words before the first comma or the first 7 words if there is no comma. 
The model structure is described in Figure~\ref{fig:salutationmodel}. It includes a word embedding layer, 
a convolutional block followed by a dropout layer, 
and two fully connected layers with 64 units and 2 units respectively.

\subsection{Full Model}

Finally, our full model combines the above four sub-models. There are multiple options to combine these above models:
\begin{itemize}\setlength\itemsep{-0.3em}
    \item at raw feature level: use all input features from each sub-model and train a classifier from scratch;
    \item at representation level: use learnt representations before the softmax layer from each sub-model;
    \item at final output level: use the predicted scores directly as the input of the final model;
    \item at final output level with rectification: use the predicted scores with rectification as the input of the final model. 
\end{itemize}

Before building separate sub-models as proposed, we experiment the first option: creating one model using all
raw features (subject, content, sender email address and name) as input, and training all parameters all at once. 
This resulting model turns out to perform slightly worse than the content model using subject and content only, 
due to the over-fitting issue brought by sender email address. 
Comparing the last two options ``at final output level directly versus with rectification'', our experiments
show that using rectification produce a better performance. 
Using the predicted scores without any correction from sender model and action model 
make the model confused for the cases when the predicted scores are not confident (not close to 1.0 or 0.0).
Thus, to avoid the confusion brought by predictions with low-confidence, 
we use the final output with rectification in order to effectively make use of the confident predictions by sub-models. 
Let $f(p, q)$ denote a ReLU activation function with modification, where $p$ is the predicted probability score output. 
For a threshold $q$($0 \leq q \leq 1$), $f(p, q) = p$ if $p \geq q$ otherwise $f(p, q) = 0$. 
In this work $q=0.99$ is used. 
Let $p$ be the estimated probability score for a label, we use $p^{+} = f(p, 0.99)$ and 
$p^{-} = f(1-p, 0.99)$ as final output with rectification
to capture the strong positive and negative signal from the model output. 
For better intuition, for examples, 
fixing $q$ at 0.99, $p=0.995$ will result in $p^{+}=0.995$ and $p^{-}=0.0$; 
$p=0.005$ will result in 0.0 and 0.995 correspondingly; 
a probability score between 0.01 and 0.99 for $p$ will produce 0.0 for both positive and negative signals.

\begin{figure*}[!t]  
  \centering
  \includegraphics[width=0.8\textwidth]{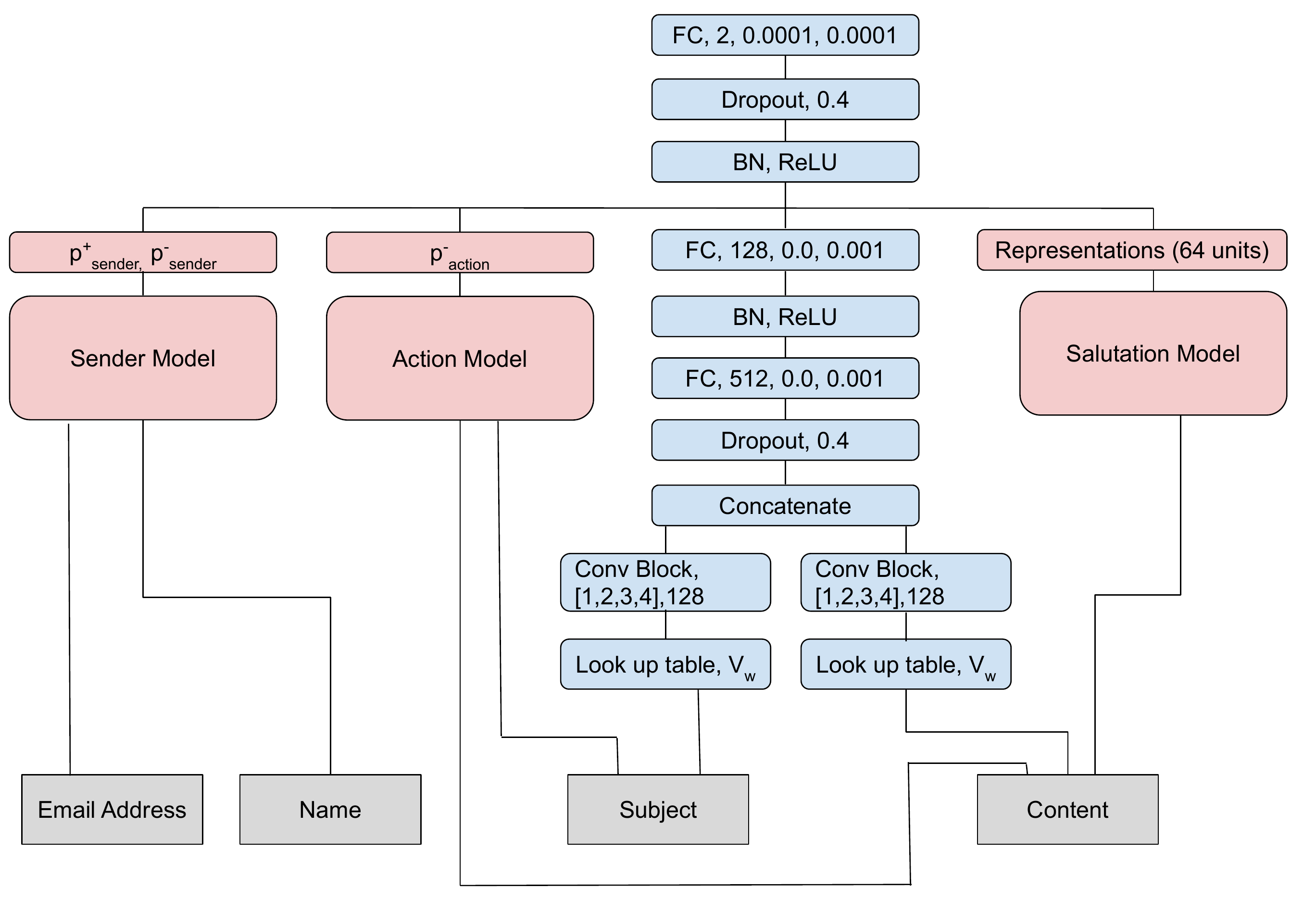}
  \caption{Model structure for full model.}
  \label{fig:fullmodel}
\end{figure*}

After experimenting with different options, the final full model (Figure~\ref{fig:fullmodel}) uses the content model as base model, 
and combines (1) final predicted probability scores with rectification from both sender model and action model, 
(2) a lower-level representation layer from salutation model.
In detail, the following items are concatenated to build the final classifier:
\begin{itemize}\setlength\itemsep{-0.3em}
    \item content model: representations with 128 hidden units, trainable for the full model; 
    \item sender model: positive and negative signals denoted by $p^{+}_{sender}$ and $p^{-}_{sender}$, 
    where $p_{sender}$ is estimated probability for positive label from sender model's final output;
    \item action model: negative signals denoted by $p^{-}_{action}$,
    where $p_{action}$ is estimated probability for positive label from action model's final output;
    \item salutation model: representations with 64 hidden units before the final layer.
\end{itemize}

In this full model, all parameters from the base model are trainable, 
while all outputs from the other three models are fixed by freezing their graphs. 
As shown in Figure~\ref{fig:fullmodel}, 
boxes in light red color represents fixed model or fixed parameters that are not trainable in the full model. 
As mentioned before, we do not train all parameters at once
in the final model to avoid over-fitting by some features. 
In contrast, all parameters in the base model (in blue) are trainable, to fine-tune
the full model. The model performance results will be shown in Section~\ref{sec:experiment}. 

\section{Data}
\label{sec:data}

Before describing the experiments setup, we first describe 
how training data and testing data are collected.

\subsection{Data Labeling}

Both the training and testing data are constructed by sampling from the Yahoo Mail corpus.
During this process, user privacy was protected through a strict data policy.
Data storage and retention follow Verizon Media policies for General Data Protection Regulation (GDPR) enforcement.
All people involved with this project had limited access to messages in compliance with Verizon Media privacy policies.

In this work, we consider three types of labeling data: manual labeling from human editors,
inheriting labels from previous Yahoo researchers' work \citep{Grbo:Hala:how:2014}, creating pseudo labels
from old production model. 

\paragraph{Manual Labeling} 
Manual labeling consists of having human editors assign labels to small set of email messages.
Given  mail data sensitivity, only a very small number of users have agreed to have their emails examined, and only a limited number of responsible editors are involved.

Different sets of email data for editorial judgements were created based on two techniques
- active learning technique and over-sampling technique.
Active learning was used to create hard examples iteratively, to help model training.
Over-sampling was used to create validation data (4K+ messages) and test data (4K+ messages), for resolving the small class issue (details presented in Appendix). Note that the data sampled by active learning
was used for model training only. All test/evaluation data were obtained via over-sampling strategy. For our human/machine classification problem, human class is a minor class. 
To alleviate this imbalanced data issue, 
we proposed to over-sample positive data from model-predicted positive data to obtain test data; this can greatly improve the confidence for precision metrics without hurting recall metrics. This effect of oversampling on test data was analyzed analytically in a probabilistic framework; this piece of work was submitted to another conference recently.

\paragraph{Inheriting Labels} 
Previous Yahoo researchers \citep{Grbo:Hala:how:2014} took great effort to collect ground-truth labels for human and machine category at sender level.
Their work left us a total around 200K senders, with approximately 100K senders labeled as human and machine each. 

\paragraph{Creating Pseudo Labels} 
We created pseudo labels by taking advantage of the great performance of the previous production model.
Senders with very confident prediction scores are sampled as pseudo labels. The number of pseudo labels
created from this source will be described in Section~\ref{sec:data_for_models} for each model separately.

\subsection{Training Data Sampling}
\label{sec:data_for_models}

Despite drawing labeled messages based on senders from inheriting labels and creating pseudo labels,
the distribution is skewed towards higher volume senders (machine senders in most cases). 
As a result, training models on skewed data often lead to over-fitting and causes biases
towards larger senders. We alleviate this issue by (1) removing duplicated messages with the same sender and subject within each day;
(2) limiting the number of samples within one day for both human and machine senders.

For content model, most of training data are obtained by sampling messages from labeled senders: 
200K senders from inherited labels and 1M human senders from creating pseudo labels. 
Note that pseudo labels are created for human senders only to obtain more balanced data at message level. 
Sampling one month's data produces around 9M samples. 
The messages labeled by editors based on active learning 
are also included with duplicating them (10$\times$ - 50$\times$).
The same training data is also used for the full model. 
For sender model, we sampled 2M positive and 2M negative pseudo labels at sender level. 
The training data for the action model are sampled based on email recipients' open/deleting actions. We sampled recent three days of data, resulting in around 8M messages.
The training data for the salutation model is the same as the training data for the content model, but each message was re-labeled based on whether it contains ``explicit salutation'' (labeled as 1) or not (labeled as 0).

\section{Experiments}
\label{sec:experiment}

To illustrate the effectiveness of the full model architecture, we compare and present the results of our models against the strong baseline in the old production model. We also evaluate the performance of the Bert model to see how our full model (Figure~\ref{fig:fullmodel}) performs compared against the state-of-the-art model in this H/M task. Different maximum sequence length for content $s_{content}$ during inference time are also evaluated for content model.

\subsection{Experiment Configuration}
\label{sec:experi_config}

As discussed in Section~\ref{sec:model}, the four sub-models are trained separately before combining them.
All these models use Adam optimizer with learning rate 0.001 and batch size 128.
We use dropout rate 0.4, 0.6, 0.4, 0.6 for content, sender, action, and salutation model respectively. 
For deciding maximum sequence length $s$ during training, we use $s_{subject}=30$ and $s_{content}=1000$
for content model and action model; use $s_{address}=1000$ (representing the number of characters)
and $s_{name}=30$ for sender model; use $s_{salutation}=10$ for salutation model. 
During inference, we change $s_{content}$ from 1000 to 2000, by which only the recall metric can improve.   
Note that all sequence lengths are at word level, except for $s_{address}$ at character level. 

\paragraph{Vocabulary Dictionary Selection}
Individual dictionary is built based on each model's specific needs.
Content model and action model share the same vocabulary dictionary;
we choose top 400K words based on frequency and top 400K words based on chi-square statistics, 
producing 598,378 unique words in total for $V_{w}$.
Similarly, for salutation model, top 200K words based on frequency and top 200K words based on chi-square statistics
are selected, producing 346,292 unique words in total for $V_{salutation}$.
For sender model, two separate dictionaries $V_{trig}$ and $V_{name}$ are built for sender address's letter-trigram
and sender name. The 30K top-frequency letter-trigrams construct $V_{trig}$, 
and the 200K top-frequency words construct $V_{name}$. 

We use embedding dimension 64 for content, sender, and action model;
and dimension 128 for salutation model. 
We experiment training content model with our own dictionary in contrast to using GloVe pre-trained word embedding.
With our own dictionary with embedding dimension 64, we achieve the same model performance
as GloVe with embedding dimension 300. 
This reduces our model size by almost 4 times compared to using GloVe embedding.

\subsection{Experiment Metrics}
Model performance is often measured by average precision (AP). Yet this metric is inadequate for us
since we are only interested when precision is above 90\%. 
Thus, recall at a fixed level of high precision (R@P) is a more appropriate metric.
Yet, due to the fact that human class is a very small class (accounting for only 3-5\% of all messages), test data produced by random sampling will contain a very small number of positive-class samples since resources for labeling test data are limited. 
With a small number of samples for positive class, we are not able to obtain accurate precision/recall metrics.
To overcome this issue, 
we over-sample the small class and adjust the corresponding precision and recall
afterwards such that it can reflect the real-world scenario. Please see details in Appendix. We use adjusted-recall, fixing adjusted-precision at 90\% and 96\% individually (Adj-R@P=90\% and Adj-R@P=96\%) as our performance metrics.  

\subsection{Experimental Results}

\begin{table}[!t]
\centering
\small
\scalebox{1.0}{
\begin{tabular}{rrrr}
\hline
Model & $s_{content}$ & Adj-R & Adj-R \\
   &  & @P=90\% & @P=96\%  \\
\hline
Bert(pretrained) & 512   & 78.5\% & 70.5\%   \\
Content(CNN)      & 512   & 77.8\% & 69.5\%   \\
Content(CNN)      & 2000  & 81.1\% & 70.4\%   \\
Full Model         & 2000  & 83.6\% & 78.8\%   \\
\hline
\end{tabular}
}
\caption{Adj-R@P=90\% and Adj-R@P=96\% performance metrics for the Bert model, content model with $s_{content}$=512 and 2000 separately, and full model for H/M classification tasks.}
\label{tab:results}
\end{table}

The experimental results on the editorial test data are summarized in Table~\ref{tab:results}. For all models we trained (including the Bert model), the best checkpoint is
selected based on our validation data. The test data is used only for model inference. Results show that our full model gives the best overall performance for both metrics Adj-R@P at 90\% and 96\%.
For reference, the old production model based on logistic regression produces adjusted-precision 94.7\% and adjusted-recall 70.5\%. We do not present the corresponding two metrics Adj-R@P=90\% and Adj-R@P=95\% in Table~\ref{tab:results} like we do for the other models, since the predicted probability scores for the old production model are not obtainable.
Comparing our content model with $s_{content}=2000$ against the old production model, it produces adj-recall 70.4\% at adj-precision 96\%, which is better than adj-recall 70.5\% at adj-precision 94.7\%. 
The full model, which combines the sender model, the action model, and the salutation model on top of the content model, provides huge performance gains by improving Adj-R@P=96\% from 70.4\% to 78.8\% (a relative 11.9\% gain) and Adj-R@P=90\% from 81.1\% to 83.6\% (a relative 3.1\% gain). 
This performance boost comes from utilizing the sender feature and the two behavior signals:
senders' explicit salutation signal and recipients' opening/deleting behaviors.
With the full model improving the adj-recall from 70.5\% to 78.8\% over the old production model (and with the adj-precision increasing from 94.7\% to 96.0\%),
this results in a 28\% reduction in the number of missed personal emails, compared to the old production model.
Note that we also experimented different combinations of sub models. 
Each sub-model itself is performing reasonably well for this binary classification task; 
yet combining the content model and one another sub model does not always perform significantly better than the content model itself. We found that the sub models interact with each other so that the final model sees bigger improvement than the aggregated contributions of all the sub models. The final model performs best due to some interactions among embeddings from each sub-model.

Besides comparing against the old production model, we also tried the state-of-the-art Bert model for this task. We experimented with two variants in addition to hyperparameters tuning: (1) fine-tuning only for our task using the Bert published model as the initial checkpoint; (2) pre-training for 200K steps over mail data using the Bert published model as the initial model and then fine-tuning for our task afterwards. We found that pre-training over mail data before fine-tuning performs significantly better for the task than fine-tuning only. To our surprise, the Bert model performed worse than even our content model with $s_{content}=2000$. This may be explained by the different maximum number of text input for the Bert model versus our content model with $s_{content}=2000$.
Due to the restriction of maximum text length in the published Bert model, we can only use at most 512 as the maximum text length for the Bert model, because of the high memory requirement for the self-attention part.
Sometimes the latter part of an email can include important signals for classifying H/M messages, for example ``subscribe'' and ``unsubscribe''. To make the comparison more even, we calculated performance metrics for the content model with $s_{content}=512$, and found that the Bert model performs only 1\% better at the adjusted-recall. 

\section{Discussion and Future Work}
\label{sec:conclude}

In this paper we present a high-quality classification algorithm for H/M classification task. 
We built and trained individual CNN models:
(1) a \textit{content model} with subject and content as input; 
(2) a \textit{sender model} with sender email address and name as input;
(3) an \textit{action model} by analyzing email recipients' action patterns on human/machine messages and 
correspondingly generating target labels based on senders' opening/deleting behaviors;
(4) a \textit{salutation model} by taking advantage of senders' ``explicit salutation'' signal as positive labels.
Then we combined the above four models into a final \textit{full model} after exploring potential options for combination: (a) at raw feature level; (b) at representation level; (c) at final output level; (d) at final output level with rectification. In this work, we find out 
that option (b) is better than (a) due to the over-fitting issue by sender email address feature; 
option (d) is better than (c) for cases with low-confident prediction.  
Experimental results show that our full model gives the best overall performance for both metrics Adj-R@P at 90\% and 96\%, with a significant improvement than the old production model. Our full model also outperforms the state-of-the art Bert model significantly at this task. 
The full model was deployed to current production system and its predictions are updated on a daily basis. 
We launched a new feature called ``People view'' in Yahoo! Mail, which shows personal communications and is one of the most loved features.
Our future work includes: (1) improve the model performance by leveraging the transformer architecture; (2) extend our model to international market.

\section*{Acknowledgments}
We are grateful to the Mail engineering team, data team, and product team at Yahoo for 
working closely and spending efforts on deploying this model into production. 
This work would not be possible without their collaboration. 
We would like to thank the annotator team for spending a huge number of hours judging emails.
We would also like to thank Dr. Alyssa Glass for the direction at the early stage of this project
and Dr. Hong Zhang for the valuable contribution at the early stage of this project.

\bibliography{aaai22}

\begin{thebibliography}{29}
\providecommand{\natexlab}[1]{#1}

\bibitem[{Agarwal and Singh(2018)}]{Agar:Jite:Temp:2018}
Agarwal, M.~K.; and Singh, J. 2018.
\newblock Template Trees: Extracting Actionable Information from Machine
  Generated Emails.
\newblock In \emph{International Conference on Database and Expert Systems
  Applications}, 3--18. Springer.

\bibitem[{Ailon et~al.(2013)Ailon, Karnin, Liberty, and
  Maarek}]{Ailo:Karn:Thre:2013}
Ailon, N.; Karnin, Z.~S.; Liberty, E.; and Maarek, Y. 2013.
\newblock Threading machine generated email.
\newblock In \emph{Proceedings of the sixth ACM international conference on Web
  search and data mining}, 405--414. ACM.

\bibitem[{Azarbonyad, Sim, and White(2019)}]{Azar:Sim:Doma:2019}
Azarbonyad, H.; Sim, R.; and White, R.~W. 2019.
\newblock Domain adaptation for commitment detection in email.
\newblock In \emph{Proceedings of the Twelfth ACM International Conference on
  Web Search and Data Mining}, 672--680. ACM.

\bibitem[{Bojanowski et~al.(2017)Bojanowski, Grave, Joulin, and
  Mikolov}]{Boja:Grav:Enri:2017}
Bojanowski, P.; Grave, E.; Joulin, A.; and Mikolov, T. 2017.
\newblock Enriching word vectors with subword information.
\newblock \emph{Transactions of the Association for Computational Linguistics},
  5: 135--146.

\bibitem[{Devlin et~al.(2018)Devlin, Chang, Lee, and
  Toutanova}]{Devl:Chan:Bert:2018}
Devlin, J.; Chang, M.-W.; Lee, K.; and Toutanova, K. 2018.
\newblock Bert: Pre-training of deep bidirectional transformers for language
  understanding.
\newblock \emph{arXiv preprint arXiv:1810.04805}.

\bibitem[{Di~Castro et~al.(2018)Di~Castro, Gamzu, Grabovitch-Zuyev,
  Lewin-Eytan, Pundir, Sahoo, and Viderman}]{Di:Ifta:Auto:2018}
Di~Castro, D.; Gamzu, I.; Grabovitch-Zuyev, I.; Lewin-Eytan, L.; Pundir, A.;
  Sahoo, N.~R.; and Viderman, M. 2018.
\newblock Automated Extractions for Machine Generated Mail.
\newblock In \emph{Companion Proceedings of the The Web Conference 2018},
  655--662. International World Wide Web Conferences Steering Committee.

\bibitem[{Di~Castro et~al.(2016)Di~Castro, Karnin, Lewin-Eytan, and
  Maarek}]{Cast:Zoha:You:2016}
Di~Castro, D.; Karnin, Z.; Lewin-Eytan, L.; and Maarek, Y. 2016.
\newblock You've got mail, and here is what you could do with it!: Analyzing
  and predicting actions on email messages.
\newblock In \emph{Proceedings of the Ninth ACM International Conference on Web
  Search and Data Mining}, 307--316. ACM.

\bibitem[{Douzi et~al.(2020)Douzi, AlShahwan, Lemoudden, and
  Ouahidi}]{douzi2020hybrid}
Douzi, S.; AlShahwan, F.~A.; Lemoudden, M.; and Ouahidi, B. 2020.
\newblock Hybrid email spam detection model using artificial intelligence.
\newblock \emph{International Journal of Machine Learning and Computing},
  10(2): 316--322.

\bibitem[{Edizel, Mantrach, and Bai(2017)}]{Ediz:Mant:Deep:2017}
Edizel, B.; Mantrach, A.; and Bai, X. 2017.
\newblock Deep character-level click-through rate prediction for sponsored
  search.
\newblock In \emph{Proceedings of the 40th International ACM SIGIR Conference
  on Research and Development in Information Retrieval}, 305--314. ACM.

\bibitem[{Grbovic et~al.(2014)Grbovic, Halawi, Karnin, and
  Maarek}]{Grbo:Hala:how:2014}
Grbovic, M.; Halawi, G.; Karnin, Z.; and Maarek, Y. 2014.
\newblock How many folders do you really need?: Classifying email into a
  handful of categories.
\newblock In \emph{Proceedings of the 23rd ACM International Conference on
  Conference on Information and Knowledge Management}, 869--878. ACM.

\bibitem[{Ioffe and Szegedy(2015)}]{Ioff:Szeg:batc:2015}
Ioffe, S.; and Szegedy, C. 2015.
\newblock Batch normalization: Accelerating deep network training by reducing
  internal covariate shift.
\newblock 448--456.

\bibitem[{Joulin et~al.(2016)Joulin, Grave, Bojanowski, and
  Mikolov}]{Joul:Grav:Bag:2016}
Joulin, A.; Grave, E.; Bojanowski, P.; and Mikolov, T. 2016.
\newblock Bag of tricks for efficient text classification.
\newblock \emph{arXiv preprint arXiv:1607.01759}.

\bibitem[{Kiros et~al.(2015)Kiros, Zhu, Salakhutdinov, Zemel, Urtasun,
  Torralba, and Fidler}]{Kiro:Zhu:Skip:2015}
Kiros, R.; Zhu, Y.; Salakhutdinov, R.~R.; Zemel, R.; Urtasun, R.; Torralba, A.;
  and Fidler, S. 2015.
\newblock Skip-thought vectors.
\newblock In \emph{Advances in neural information processing systems},
  3294--3302.

\bibitem[{Kocayusufoglu et~al.(2019)Kocayusufoglu, Sheng, Vo, Wendt, Zhao,
  Tata, and Najork}]{Koca:Shen:Rise:2019}
Kocayusufoglu, F.; Sheng, Y.; Vo, N.; Wendt, J.; Zhao, Q.; Tata, S.; and
  Najork, M. 2019.
\newblock {RiSER}: Learning Better Representations for Richly Structured
  Emails.
\newblock In \emph{The World Wide Web Conference}, 886--895. ACM.

\bibitem[{Kumaresan, Saravanakumar, and
  Balamurugan(2019)}]{Kuma:Sara:Visu:2019}
Kumaresan, T.; Saravanakumar, S.; and Balamurugan, R. 2019.
\newblock Visual and textual features based email spam classification using
  S-Cuckoo search and hybrid kernel support vector machine.
\newblock \emph{Cluster Computing}, 22(1): 33--46.

\bibitem[{Lan et~al.(2020)Lan, Chen, Goodman, Gimpel, Sharma, and
  Soricut}]{lan2019albert}
Lan, Z.; Chen, M.; Goodman, S.; Gimpel, K.; Sharma, P.; and Soricut, R. 2020.
\newblock Albert: A lite bert for self-supervised learning of language
  representations.
\newblock In \emph{The International Conference on Learning Representations}.

\bibitem[{Le and Mikolov(2014)}]{Le:Miko:Dist:2014}
Le, Q.; and Mikolov, T. 2014.
\newblock Distributed representations of sentences and documents.
\newblock In \emph{International conference on machine learning}, 1188--1196.

\bibitem[{Liu et~al.(2019)Liu, Ott, Goyal, Du, Joshi, Chen, Levy, Lewis,
  Zettlemoyer, and Stoyanov}]{Liu:Ott:Robe:2019}
Liu, Y.; Ott, M.; Goyal, N.; Du, J.; Joshi, M.; Chen, D.; Levy, O.; Lewis, M.;
  Zettlemoyer, L.; and Stoyanov, V. 2019.
\newblock Roberta: A robustly optimized bert pretraining approach.
\newblock \emph{arXiv preprint arXiv:1907.11692}.

\bibitem[{Mikolov et~al.(2013)Mikolov, Sutskever, Chen, Corrado, and
  Dean}]{Miko:Suts:Dist:2013}
Mikolov, T.; Sutskever, I.; Chen, K.; Corrado, G.~S.; and Dean, J. 2013.
\newblock Distributed representations of words and phrases and their
  compositionality.
\newblock In \emph{Advances in neural information processing systems},
  3111--3119.

\bibitem[{Mohammadzadeh and Gharehchopogh(2021)}]{mohammadzadeh2021novel}
Mohammadzadeh, H.; and Gharehchopogh, F.~S. 2021.
\newblock A novel hybrid whale optimization algorithm with flower pollination
  algorithm for feature selection: Case study Email spam detection.
\newblock \emph{Computational Intelligence}, 37(1): 176--209.

\bibitem[{Potti et~al.(2018)Potti, Wendt, Zhao, Tata, and
  Najork}]{Pott:Wend:Hidd:2018}
Potti, N.; Wendt, J.~B.; Zhao, Q.; Tata, S.; and Najork, M. 2018.
\newblock Hidden in plain sight: Classifying emails using embedded image
  contents.
\newblock In \emph{Proceedings of the 2018 World Wide Web Conference},
  1865--1874. International World Wide Web Conferences Steering Committee.

\bibitem[{Sheng et~al.(2018)Sheng, Tata, Wendt, Xie, Zhao, and
  Najork}]{Shen:Tata:Anat:2018}
Sheng, Y.; Tata, S.; Wendt, J.~B.; Xie, J.; Zhao, Q.; and Najork, M. 2018.
\newblock Anatomy of a privacy-safe large-scale information extraction system
  over email.
\newblock In \emph{Proceedings of the 24th ACM SIGKDD International Conference
  on Knowledge Discovery \& Data Mining}, 734--743. ACM.

\bibitem[{Shu et~al.(2020)Shu, Mukherjee, Zheng, Awadallah, Shokouhi, and
  Dumais}]{10.1145/3397271.3401121}
Shu, K.; Mukherjee, S.; Zheng, G.; Awadallah, A.~H.; Shokouhi, M.; and Dumais,
  S. 2020.
\newblock \emph{Learning with Weak Supervision for Email Intent Detection},
  1051–1060.
\newblock New York, NY, USA: Association for Computing Machinery.
\newblock ISBN 9781450380164.

\bibitem[{Sun et~al.(2018)Sun, Garcia-Pueyo, Wendt, Najork, and
  Broder}]{Sun:Garc:Lear:2018}
Sun, Y.; Garcia-Pueyo, L.; Wendt, J.~B.; Najork, M.; and Broder, A. 2018.
\newblock Learning Effective Embeddings for Machine Generated Emails with
  Applications to Email Category Prediction.
\newblock In \emph{2018 IEEE International Conference on Big Data (Big Data)},
  1846--1855. IEEE.

\bibitem[{Vaswani et~al.(2017)Vaswani, Shazeer, Parmar, Uszkoreit, Jones,
  Gomez, Kaiser, and Polosukhin}]{Vasw:Shaz:Atte:2017}
Vaswani, A.; Shazeer, N.; Parmar, N.; Uszkoreit, J.; Jones, L.; Gomez, A.~N.;
  Kaiser, {\L}.; and Polosukhin, I. 2017.
\newblock Attention is all you need.
\newblock In \emph{Advances in neural information processing systems},
  5998--6008.

\bibitem[{Vinayakumar et~al.(2019)Vinayakumar, Soman, Poornachandran, Mohan,
  and Kumar}]{Vina:Soma:Scal:2019}
Vinayakumar, R.; Soman, K.; Poornachandran, P.; Mohan, V.~S.; and Kumar, A.~D.
  2019.
\newblock ScaleNet: scalable and hybrid framework for cyber threat situational
  awareness based on DNS, URL, and email data analysis.
\newblock \emph{Journal of Cyber Security and Mobility}, 8(2): 189--240.

\bibitem[{Whittaker et~al.(2019)Whittaker, Edmonds, Tata, Wendt, and
  Najork}]{Whit:Edmo:Onli:2019}
Whittaker, M.; Edmonds, N.; Tata, S.; Wendt, J.~B.; and Najork, M. 2019.
\newblock Online template induction for machine-generated emails.
\newblock \emph{Proceedings of the VLDB Endowment}, 12(11): 1235--1248.

\bibitem[{Yang et~al.(2019)Yang, Dai, Yang, Carbonell, Salakhutdinov, and
  Le}]{Yang:Dai:XLNe:2019}
Yang, Z.; Dai, Z.; Yang, Y.; Carbonell, J.; Salakhutdinov, R.; and Le, Q.~V.
  2019.
\newblock XLNet: Generalized Autoregressive Pretraining for Language
  Understanding.
\newblock \emph{arXiv preprint arXiv:1906.08237}.

\bibitem[{Yang et~al.(2016)Yang, Yang, Dyer, He, Smola, and
  Hovy}]{Yang:Yang:Hier:2016}
Yang, Z.; Yang, D.; Dyer, C.; He, X.; Smola, A.; and Hovy, E. 2016.
\newblock Hierarchical attention networks for document classification.
\newblock In \emph{Proceedings of the 2016 conference of the North American
  chapter of the association for computational linguistics: human language
  technologies}, 1480--1489.

\end{thebibliography}
\bibliographystyle{aaai22}

\clearpage
\section*{Appendix}
\label{sec:append}

\paragraph{The Annotation Guideline Details} 

It is not straightforward to know whether an email was technically sent by a human or a machine. The guideline is about the nature of the content of the message rather than who technically sent it. If the content of the message is ``personal'' (communicated within a small group of people who know one another), then it is labeled as ``Human''. Otherwise, it will be labeled as ``Machine''. This definition is well aligned with an alternative definition (based on who technically sent it) even though there can be exceptions. Every message in the test data is judged by at least two annotators. 
For each message, if all annotators make the same judgement, 
then this judgement will be used as the label of this message.
Or else, an additional annotator will be assigned to make the final decision.

\paragraph{Intuition for Adjusting Precision/Recall}

Before we go into the notations, we would like to explain why this adjustment is the right way to get correct adjusted precision/recall numbers. For a better intuition, we assume (in this Intuition section only) that the model used for sampling is the same as the model we would like to calculate performance metrics on.
Let's start with the whole population and a classifier, 
which gives the following confusion matrix in Table~\ref{tab:ConfMatrixP}. If we haven't done any editorial test, we don't know the values 
of $n_{1,1}$, $n_{1,0}$, $n_{0,1}$, and $n_{0,0}$. 
Those are unknown but fixed values. 
Note that precision is $n_{1,1}/(n_{1,1}+n_{0,1})$ and
recall is $n_{1,1}/(n_{1,1}+n_{1,0})$.

\begin{table}[ht]
\centering
\small
\begin{tabular}{rrr}
\hline
 & Predicted$+$ & Predicted$-$ \\
\hline
Actual$+$ & $n_{1,1}$ & $n_{1,0}$     \\
Actual$-$ & $n_{0,1}$ & $n_{0,0}$     \\
\hline
\end{tabular}
\caption{Confusion Matrix: Population}
\label{tab:ConfMatrixP}
\end{table}

Now we randomly sample examples from the predicted$+$ group and the predicted$-$ group using two ratios r and r' respectively and actually do an editorial test, which will yield the confusion matrix in Table~\ref{tab:ConfMatrixS}. 
Note that the numbers in Table~\ref{tab:ConfMatrixS} are the expectations if we do sampling many times.

\begin{table}[ht]
\centering
\small
\begin{tabular}{rrr}
\hline
 & Predicted$+$ & Predicted$-$ \\
\hline
Actual$+$ & $n_{1,1} \cdot r$ & $n_{1,0} \cdot r'$     \\
Actual$-$ & $n_{0,1} \cdot r$ & $n_{0,0} \cdot r'$     \\
\hline
\end{tabular}
\caption{Confusion Matrix: Sampling}
\label{tab:ConfMatrixS}
\end{table}

Since we did an editorial test, we now know the values of $n_{1,1} \cdot r$, $n_{1,0} \cdot r'$, $n_{0,1} \cdot r$ and $n_{0,0} \cdot r'$. Note that precision is still $n_{1,1}/(n_{1,1}+n_{0,1})$ but recall becomes $n_{1,1} \cdot r/(n_{1,1} \cdot r+n_{1,0} \cdot r')$. Then, we need to adjust the last column in the matrix by multiplying $r/r'$ and get the following confusion matrix in Table~\ref{tab:ConfMatrixAdj}.
\begin{table}[ht]
\centering
\small
\begin{tabular}{rrr}
\hline
 & Predicted$+$ & Predicted$-$ \\
\hline
Actual$+$ & $n_{1,1} \cdot r$ & $n_{1,0} \cdot r'\cdot (r/r')$     \\
Actual$-$ & $n_{0,1} \cdot r$ & $n_{0,0} \cdot r'\cdot (r/r')$     \\
\hline
\end{tabular}
\caption{Confusion Matrix: Adjustment}
\label{tab:ConfMatrixAdj}
\end{table}

After this adjustment , we can get the same precision and recall values as the whole population.

\paragraph{The Adjusted Precision/Recall} 

The section illustrates
how we did over-sampling and adjusted precision/recall for the small class ($+$ class).

We did over-sampling based on a model $s$ (denoted by $\psi_s$); 
then calculated performance metrics on a model $f$ (denoted by $\psi_f$). 
First a large number of data are randomly sampled from the population. 
Model $\psi_s$ splits this random sample into two groups $G^{+}$ and $G^{-}$ 
by assigning data with probability score above a cut-off threshold into group $G^{+}$ 
and the remaining in group $G^{-}$, 
resulting in a group size ratio $k=G^{-}/G^{+}$.
Next, we randomly sample $M_{s+}$ and $M_{s-}$ data points
from $G^{+}$ and $G^{-}$ respectively. 
These ($M_{s+}+M_{s-}$) samples will be used as test data and judged by editors.
Assuming $P$ and $N$ of these samples are manual-labeled as positive and negative labels respectively ($M_{s+}+M_{s-}=P+N$). 
For adjusting precision/recall metrics, the $M_{s-}$ data points should be duplicated
$\frac{M_{s+}}{M_{s-}} \cdot k$ (denoted by $\beta$) times to recover the true $+$/$-$ proportion in real scenario. 
Table~\ref{tab:oversample} shows the expanded confusion matrix
with metrics from both $\psi_s$ and $\psi_f$, where $\beta$ is multiplied for data predicted as negative by $\psi_s$
to adjust the metrics. Adjusted-precision and adjusted-recall are following:
\begin{equation}
        Prec_{(adj)}  =  \frac{P_{s+,f+}+\beta \cdot P_{s-,f+}}{(P_{s+,f+}+\beta \cdot P_{s-,f+}) + (N_{s+,f+}+\beta \cdot  N_{s-,f+})}
\end{equation}
and
\begin{equation}
        Recall_{(adj)}  = \frac{P_{s+,f+}+\beta \cdot P_{s-,f+}}{(P_{s+,f+}+\beta \cdot P_{s-,f+}) +(P_{s+,f-}+\beta \cdot  P_{s-,f-})}.
\end{equation}

\begin{table}[b]
\centering
\small
\begin{tabular}{r|rr}
\hline
 & +($\psi_f$) & -($\psi_f$)  \\
\hline
$+$(judge),$+$($\psi_s$) & $P_{s+,f+}$ & $P_{s+,f-}$  \\
$+$(judge),$-$($\psi_s$) &  $\beta \cdot P_{s-,f+}$ & $\beta \cdot P_{s-,f-}$  \\
$-$(judge),$+$($\psi_s$) & $N_{s+,f+}$ & $N_{s+,f-}$  \\
$-$(judge),$-$($\psi_s$) &  $\beta \cdot N_{s-,f+}$ & $\beta \cdot N_{s-,f-}$  \\
\hline
\end{tabular}
\caption{Adjusted expanded confusion matrix.Here, 
$P$ and $N$ represents the number of editor-labeled positive and negative cases respectively; 
subscripts $_{s+}$, $_{s-}$, $_{f+}$, $_{f-}$ correspond to being predicted as positive by $\psi_s$, as negative by $\psi_s$, as positive by $\psi_f$, and as negative by $\psi_f$ respectively. 
For example, $P_{s+,f+}$ is the number of editor-labeled positive samples predicted as positive from both $\psi_s$ and $\psi_f$, 
$N_{s+,f+}$ is the number of editor-labeled negative samples predicted as positive from both $\psi_s$ and $\psi_f$.}
\label{tab:oversample}
\end{table}

There are two specific scenarios: 
\begin{itemize}\setlength\itemsep{-0.3em}
    \item $\beta=1$: random-sampling, thus adjusted precision/recall is the same as before adjustment.
    \item $\psi_s = \psi_f$: two models are the same, so the adjusted-precision remains the same as before adjustment.
\end{itemize} 

In practice the two models are different and typical values for beta are between 1 and 10. This depends on the imbalance ratio, the estimated precision and recall, and the desired precision/recall accuracy level.

\end{document}